# Enlightening Deep Neural Networks
# with Knowledge of Confounding Factors

Yu Zhong       Gil Ettinger

{yu.zhong, gil.ettinger}@stresearch.com

Systems & Technology Research

## Abstract

*Deep learning techniques have demonstrated significant capacity in modeling some of the most challenging real world problems of high complexity. Despite the popularity of deep models, we still strive to better understand the underlying mechanism that drives their success. Motivated by observations that neurons in trained deep nets predict attributes indirectly related to the training tasks, we recognize that a deep network learns representations more general than the task at hand to disentangle impacts of multiple confounding factors governing the data, in order to isolate the effects of the concerning factors and optimize a given objective. Consequently, we propose a general framework to augment training of deep models with information on auxiliary explanatory data variables, in an effort to boost this disentanglement and train deep networks that comprehend the data interactions and distributions more accurately, and thus improve their generalizability. We incorporate information on prominent auxiliary explanatory factors of the data population into existing architectures as secondary objective/loss blocks that take inputs from hidden layers during training. Once trained, these secondary circuits can be removed to leave a model with the same architecture as the original, but more generalizable and discerning thanks to its comprehension of data interactions. Since pose is one of the most dominant confounding factors for object recognition, we apply this principle to instantiate a pose-aware deep convolutional neural network and demonstrate that auxiliary pose information indeed improves the classification accuracy in our experiments on SAR target classification tasks.*

## 1. Introduction

In recent years, deep learning, in particular deep convolutional neural networks (DCNN) [28], has taken the computer vision field by storm, setting drastically improved performance records for many real world computer vision challenges including general object recognition [7] [16][27] [38], face recognition [45], scene classification [56], object detection and segmentation [14] [20], feature encoding [25], metric learning [22], and 3D reconstruction [10]. Such dominantly superior performance by deep learning relative to other machine learning approaches has also been demonstrated in numerous other application fields – including speech recognition and natural language processing – generating unprecedented enthusiasm and optimism in artificial intelligence in both the research community and general public. This overwhelming success of deep learning is propelled by three indispensable enabling factors:

1. Groundbreaking algorithm developments in exploitation of deep architectures and effective optimization of these networks, allowing capable representation and modeling of complex problems [18][27][28];

2. Availability of very large-scale training datasets that capture full data variations in real world applications in order to train high capacity neural networks [6]; and

3. Advanced processing capability in graphical processing units (GPU) enabling computation in speed and scale that was impossible earlier.

Despite the sweeping success of DCNNs, we still strive to understand how and why they work so well in order to better utilize them. In this paper we contemplate what deep networks have actually learned once they have been trained to perform a particular task and explore how to take advantage of that knowledge. Based on observations reported in multiple research efforts that neurons in trained deep networks predict attributes that are not directly associated with the training tasks, we perceive that deep models learn structures more general than what the task at hand involves. This generalizability likely results from performance optimization on data populations with multiple explanatory factors, which is typical in real world applications. As a result, we can assist the unsupervised learning of latent factors that naturally occur in the supervised training of deep nets, by supplying information on dominant confounding factors during training. Information on such data impacting factors allows the neural network to untangle interactions in data, learn more accurate characterization of the underlying data distributions, and generalize better with new data.

In this spirit we propose to augment the training of deep convolutional neural networks using information on impacting confounding factors in order to improve their performance. We describe a general framework to boost training of any existing deep architecture with auxiliary explanatory factors in the data as long as information on such influential confounding factors is available. Such information has often been overlooked because it is deemed not directly related to the task at hand. To verify the idea we study the use of auxiliary pose information to improve the classification performance of a deep neural network. As is well known, pose is a major nuisance variable in many recognition and classification tasks. In practice, we either extract pose-invariant features or explicitly model pose effects to perform accurate recognition under various poses. Typically, to train a model for classification, only class labels are used for training. In this study we inject pose information in addition to class labels during the training to obtain a pose-aware DCNN for classification. We expect that the pose information will help the model learn more accurate representation for the data population and improve the classification accuracy. We evaluate the proposed pose-aware deep architecture on a synthetic aperture radar (SAR) automatic target recognition (ATR) task [9][13] using the Moving and Stationary Target Acquisition and Recognition (MSTAR) public dataset [36]. We demonstrate that the pose information in training indeed helps improve the classification performance.

In this paper we make the following contributions.
• We start with the DCNN fundamentals and explore how DCNNs model complex real world data typically impacted by multiple explanatory factors. We observe that during supervised learning of a certain task, a DCNN simultaneously performs unsupervised learning of latent explanatory factors, likely driven by the effort to isolate the influence of the concerning factors as a result of the optimization process. Therefore, augmenting the training process with information on major data explanatory factors allows the DCNN to capture more accurate governing structure in the data, and thus improves its generalizability. We believe we are the first to study and provide an explanation for the unsupervised learning behavior arising naturally in the training of DCNNs and direct this behavior towards improving the DCNN performance. This finding improves our understanding on how DCNNs work and opens up new opportunities to improve their performance.

• We describe a general framework to augment training of DCNNs using available information on influential confounding factors of the data population. This framework can be applied to any existing deep architecture, with a very small computational cost.

• To verify this finding we propose to train a novel pose-aware DCNN architecture to explicitly encode pose information in addition to class information during training. Note that pose information is not needed for test data. We apply the pose-aware DCNN to classify targets in synthetic aperture radar (SAR) imagery and demonstrate that the pose information injected into the DCNN during training improves classification accuracy on test data.

The remainder of the paper is organized as follows. We review related literature in Section 2 and motivate our approach in Section 3. Section 4 describes a general framework to take advantage of auxiliary explanatory factors to improve the performance of DCNNs. We introduce the pose-aware architecture in Section 5. Section 6 presents the performance analysis of the proposed approach using the public MSTAR dataset and compares it to existing published results on the same dataset. We draw conclusions in Section 7.

## 2. Literature Review

Deep neural networks have demonstrated unmatched capability to tackle very complex challenges in real world context extraction applications. Understanding the fundamentals of deep neural networks helps us to better utilize them. Here, we review techniques that improve the performances of deep networks.

The capacity of a DCNN can be increased by either expanding its breadth to have more feature maps at each layer [48], or by growing the depth of the network [44]. While a deeper and broader network possesses increased representation and abstraction power to model problems of high complexity, it demands an enormous amount of training data to offset the risk of over-fitting with hundreds of millions of model parameters. As deeper architectures become more difficult to optimize, [44] has added auxiliary classifiers at intermediate layers to encourage the gradient flow to lower layers during back-propagation and improve training performance. Recently, residual networks [17] and highway networks [39] have been proposed to effectively optimize extremely deep networks.

Nonlinearity is a major factor that enables deep networks to encode complex representations. The Rectified Linear hidden Units (ReLU) have demonstrated performance gain compared to sigmoidal nonlinearities [5]. They are also fast to compute. Variants of ReLUs have also been proposed [16][29].

Hinton et al. used "drop-out" to prevent over-fitting due to co-adaptation of feature detectors by randomly dropping a portion of feature detectors during training [18]. It is analogous to boosting or a classifier ensemble in that many of these random sub-networks are trained independently and then combined in performing classification tasks to avoid over-fitting and improve robustness. Dropout training can be considered as a form of adaptive regularization to control over-fitting [47].

An alternative regularizer is batch normalization [24]. This approach integrates normalization of batch data as a part of the model architecture and performs normalization for each training mini-batch to counter the internal covariate shift during training – i.e., the input distribution to each layer changes at each epoch. As a result it allows for more aggressive training with faster convergence.

Data augmentation improves the accuracy of the trained deep models [27]. The training set and/or testing set are augmented with transformed sample images that reflect natural variations in the data distribution. This helps both achieving robustness to the variations, and combating over-fitting. Typical transforms include flipping, position jittering, cropping, and affine or other image transforms.

Multitask learning [3] trains several related tasks in parallel with a shared representation where what is learned for each task helps learning other tasks. It is argued that extra tasks serve as an inductive bias to improve the generalization of the network. [4] used a single deep network to perform a full list of similar NLP tasks including speech tagging, parsing, name-entity recognition, language model learning, and semantic role labeling. [54] proposed to optimize facial landmark detection with related tasks such as categorical head pose estimation. Recently multitask DCNNs have been successfully used to simultaneously perform multiple tasks including depth / surface normal prediction and semantic labeling [10], object detection and segmentation [14][15], and object detection, localization, and recognition [37]. Despite the success of multitask learning, its major limitation remains to better understand the mechanism how it works and to determine what kind of tasks help each other [3].

Pose is an important issue that needs to be addressed for object recognition or classification tasks [52]. In recently published work on multi-view 3D recognition, Su et al. [40] proposed a system that employs multiple DCNNs, each dedicated to a specific viewpoint, which are then pooled across views and fused using another CNN to generate a multi-view class prediction. Their hybrid multi-view architecture improved the classification accuracy from 87.3% to 88.9% on the Princeton ModelNet dataset. Zhu et al. [57] proposed generative multi-view perceptron to learn both face identities and view representations with the main purpose to generate a full spectrum of views for an input image for improved classification.

As we will evaluate the performance of the proposed pose-aware DCNN on a SAR ATR task [35][36], we also review existing literature in this field.

SAR ATR is a challenging task that demands the most advanced machine learning techniques. Ettinger et al. [12] developed a probabilistic optimization matching approach based on a Bayesian evaluation metric to support a wide range of features including points, regions, and other composite features for SAR ATR. O'Sullivan et al. proposed a conditionally Gaussian model parameterized by target type and pose, and demonstrated robustness of the models to variations in the data [32]. Bayesian compressive sensing with scattering center features has been utilized for effective and robust target classification [53]. Park and Kim recently proposed a classifier named "modified polar mapping" classifier and demonstrated good performance under various SAR extended operating conditions [33]. Other popular machine learning techniques such as support vector machines and AdaBoost have been applied to recognize targets in SAR imagery as well [42] [55].

Morgan first used a rather basic deep convolutional neural network [30] for SAR ATR. Later, "A-ConvNets" [48] was proposed to address the issue of over-fitting by replacing the fully connected layers in conventional deep neural networks with convolutional layers of local support to scale back the number of parameters for the deep network. They aggressively augmented the training set using position jitters to about 10 times the original dataset.

## 3. Motivation

Deep convolutional neural networks distinguish themselves from traditional machine learning approaches in enabling a hierarchy of concrete to abstract feature representations. A study on performance-optimized deep hierarchical models trained for object categorization and human visual object recognition abilities indicated the trained network's intermediate and top layers are highly predictive of neural responses in the visual cortex of a human brain [49]. It has been suggested that the strength of DCNNs come from the reuse and sharing of features, which results in more compact and efficient feature representations that benefit model generalization [1][2]. For example, the same convolutional filter bank is learned for the entire image domain in a DCNN, as opposed to learning location dependent filters.

An intrigue aspect of deep convolutional networks is their remarkable transferability [11] [50]. A deep network trained on one dataset is readily applicable on a different dataset. The ImageNet model by Zeiler and Fergus [51] generalizes very well on the Caltech datasets. In other works, deep models trained to perform one task, say, object recognition, can be repurposed to significantly different tasks, such as scene classification with little effort [7][19]. This natural generic modeling capabilities across tasks have also been demonstrated in the success of several integrated deep convolutional neural networks proposed to simultaneous performing multiple tasks for various applications [10][14][15][31][37]. Such facts indicate deep networks learn feature representations more pertinent about the data population than a specific task.

Furthermore, there has been significant empirical evidence emerging in the latest research that the neurons in trained DCNNs actually encode information that is not directly related to the training objectives or tasks. Semantic segregation of neuron activations on attributes such as "indoor" and "outdoor" has occurred in deep convolutional network trained for object recognition, which has prompted application of these "DeCAF" [7] features to novel generic tasks such as scene recognition and domain adaptation with success. On the other hand, Zhou et al. [56] noticed that their DCNN trained for scene classification automatically discover object categories relevant to the scene categories, even though only scene labels were used in training. Khorrami et al. [26] observed that deep neural networks that are trained for face recognition actually learn facial actions in some of its hidden neurons. The DeepID [41][43] network trained for face identification predicts gender information even through only identity labels are used in training. These observations suggest that deep models learn not only compact and reusable feature representations of data for the tasks that they are trained on, but also something more fundamental and general in order to optimize the performance.

Data populations in real applications encompass wide ranges of variations. Some are due to the factors of concern, while others are not. It is usually impossible to address one factor in isolation without taking into consideration some of other confounding factors. It is therefore necessary to either develop invariant features or explicitly model the effects of the confounding factors that significantly impact the data. For example, for face recognition applications, a face image depends on not only the identity, but also various other nuisance factors such as pose, lighting, facial expression, age, etc. An accurate face identification system needs to factor out the effects of these confounding factors.

So, when a DCNN is trained using a large data population to perform one specific task, it may naturally organize the network to capture the intrinsic data distribution governed by multiple influential explanatory factors. That is, as a result of the objective optimization, it needs to explain away the effects of the confounding factors. This explains the observed phenomenon in which neurons that predict auxiliary attributes arise in DCNNs trained for unrelated tasks.

If unsupervised learning of latent factors naturally occurs during the training of a DCNN for a specific task, boosting the training with information on influential auxiliary variables is then expected to help in reducing the error and better capturing the underlying data distribution for increased generalization power. This motivates us to investigate deep architectures that take advantage of available information on explanatory factors for improved prediction performance.

In the following sections we describe augmented training of DCNNs using dominant confounding factors. We then instantiate a pose-aware deep network use this principle, and evaluate its performance on a SAR Target classification task.

## 4. Augmented Training Using Confounding Factors

Knowledge of auxiliary explanatory factors can be easily incorporated into the deep learning framework as separate output constraints or losses in addition to the original outputs, in a way similar to multiple outputs in multi-task deep networks [3][4]. A deep network trained this way is more comprehensive and knowledgeable.

These additional constraints during training limit the solution space of the network. The constraints introduced by a dominant confounding factor likely shape the solution space in a principled way to reflect the impact of this factor on the underlying data distribution. In another perspective, this training augmentation of major explanatory factors can be considered as a form of regularization, in a general sense that "the basic idea of all regularization method is to restrict the space of possible solutions" [34]. This regularization influences the network to more accurately capture the structure due to multiple explanatory factors and their interactions, and consequently improves the network's generalizability and performance.

Figure 1 illustrates the general framework to augment training of deep convolutional neural networks. We start with a conventional architecture, consisting of convolutional layers at the bottom, then full-connected layers, and one or more prediction blocks at the top. Although the shown baseline architecture takes the simplest linear form, it can potentially be any of the existing deep architectures.

We then introduce additional objective blocks that take input from the existing hidden layers to predict one or more auxiliary confounding variables. During training, the network optimizes a weighted sum of primary objectives for the original task, and secondary objectives reflecting additional knowledge pertinent to the data distribution. The secondary objectives can be custom designed for each factor, and their weights reflect the importance and variation of these factors, which can be application dependent. The circuits to optimize the secondary objectives are fairly small compared to the remaining network. Information of confounding factors for the training data is infused to shape the network via these secondary prediction blocks during training, so that the DCNN can explicitly learn and encode both primary variables and secondary confounding factors of the input data, with a negligible cost in memory and computational time.

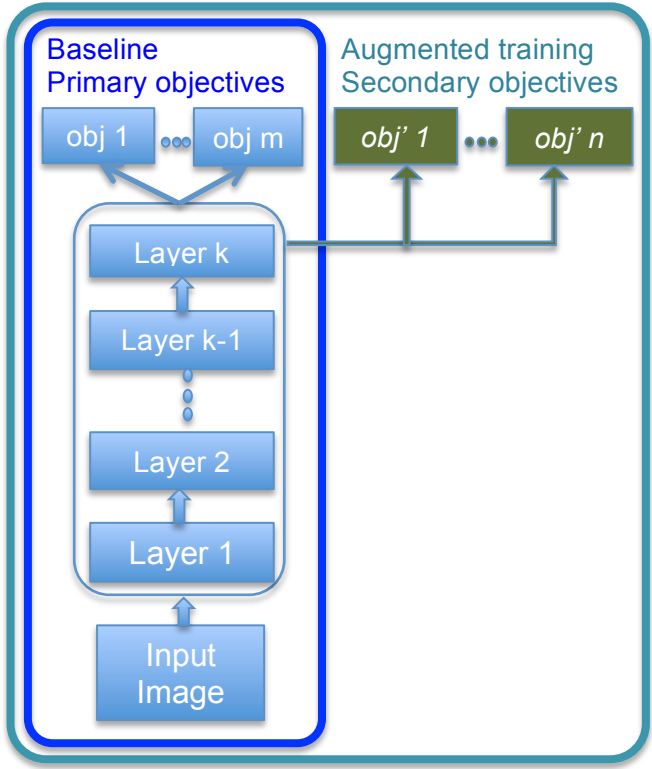

Figure 1. General framework to augment training of deep convolutional neural networks. On the left is a baseline deep convolutional neural network (shown in blue) with one or more objectives. We augment the training using influential auxiliary data explanatory variables as secondary prediction blocks (shown in green). Knowledge of these confounding variables shape the weights of the network via gradient back-propagation originated from these secondary prediction blocks.

Once the network has been trained, the auxiliary circuits for secondary factors can be removed and we obtain a model with exactly the same architecture as the baseline, in terms of number of layers and number of neurons at each layer. However, the parameter values learned with the auxiliary information make this DCNN more comprehensive and discerning to perform accurate classification during testing at no additional computational cost compared to the baseline.

Note that this general framework fully observes the principle of compact and reusable feature representation, a major strength of deep convolutional neural networks [1][2]: a single network is employed to model various factors and interactions in the input data; feature representations are shared to fulfill various objectives. We just supply the network with relevant information regarding the data and let the DCNN optimize for the best performance.

## 5. Pose-Aware Deep Convolutional Neural Networks

Pose is one of the most dominant confounding factors for recognition tasks. The same object can have drastically different appearances when the viewpoint varies. However, variations introduced by pose are systematic. Knowing the pose helps to explain away the variation induced by pose for more accurate recognition.

We investigate boosting DCNN classification performance using auxiliary pose information. For this purpose, our baseline architecture has a single prediction block at the top for classification. We introduce a secondary prediction block that takes input from the top hidden layer to regress on the pose variables. The secondary circuit introduced contains only $(N(k)+1) \times p_d$ parameters, where $N(k)$ is the number of outputs from the top hidden layer, and $p_d$ is dimension of the pose variable.

### 5.1. Objective function

We train the pose-aware DCNN using both class labels and pose information. It optimizes a linear combination of two objectives: one on the class error of the inputs (Eq **1**), and the other on the pose alignment error (Eq **3**).

To minimize the class prediction error, we use the popular softmax log-loss function for the classification task:

**Eq 1**: $obj_{class}(net, X) = - \sum_{i \in TrainSet} (x_{ic_i} - \log \sum_{j=1}^{C} e^{x_{ij}})$

where *net* stands for parameters of the network, $X$ represents the training data, $C$ is the number of classes, $x_{ij}$ is the response of the *i*th input for the *j*th class, and $x_{ic_i}$ s the response of the *i*th input for its truth class.

Different from class labels, pose variables are usually continuous. We perform linear regression using outputs from the top hidden layer to predict object pose. We only consider rotation here, as translation can be bypassed by either centering the image at object center, or by augmentation of the training data using randomly translated training images to achieve invariance to translation. In case translation needs to be explicitly modeled, it is straightforward to add it into the formulation as well.

We represent 3D rotations using quaternions [21] for their desirable properties compared to Euler angles or rotation matrices. We compute the distance between two rotations $q_1$ and $q_2$ as follows [23]:

**Eq 2:** $dist(q_1, q_2) = \arccos(|q_1 \cdot q_2|)$

where the range is $[0, \frac{\pi}{2}]$. This distance function is pseudo-metric on the unit quaternion but is a metric function on the special orthogonal group $SO(3)$ of orthogonal matrices with determinant 1. We choose it because of its fast convergence to zero compared to some of the other metric functions for rotation. It is also efficient to compute.

The loss function for pose regression is the sum of distance between rotation predicted for each training image and its truth rotation:

**Eq 3:** $obj_{pose}(net, X) = \sum_{i \in TrainSet} dist(\hat{q}_i, \tilde{q}_i)$

where $\hat{q}_i$ is the quaternion for the predicted pose of the *i*th training input and $\tilde{q}_i$ is quaternion of the associated truth pose.

During training, the following combined objective function is minimized to learn the parameters of a deep network that simultaneously predicts the class label and object pose of an input.

**Eq 4:** $obj_{combo}(net, X) = obj_{class}(net, X) + \lambda \cdot obj_{pose}(net, X)$

## 5.2. Training

During training, the combo loss function in Eq **4** is optimized so that the network learns both the category and pose information of the training inputs. Even though the ultimate goal of the network is to perform categorization, the auxiliary pose information helps the network to disentangle confounding factors that influence the input data and better characterize the categorical traits.

We follow common practices [27][46] to train the network once the objective function is defined. The weights and parameters of the network are randomly initialized with zero mean Gaussian distributions. Mini-batch Stochastic Gradient Descent is used in conjunction with momentum and weight decay.

## 5.3. Testing

The DCNN is trained with both class and pose information of the input images. Once trained, the model captures both the pose and class information and is able to predict both for a novel image. However, if pose is not the concern of the task, the sub-circuit that predicts the object pose can be removed from the network.

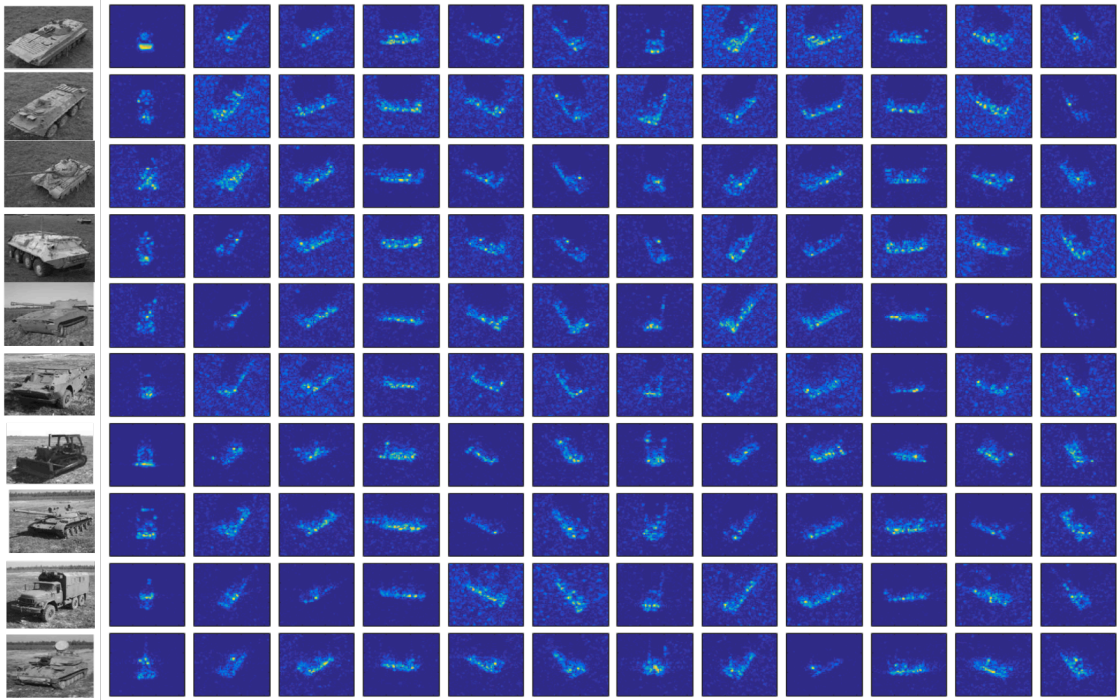

**Figure 2. Targets in the public MSTAR SAR dataset. Each row shows the picture of a target, followed by SAR chips for this target class, in the order of increasing azimuth angle, from 0º to 360º. It is evident that azimuth angle drastically affects the appearance of the SAR chips. The ten target classes are bmp2, btr70, t72, btr60, 2S1, brdm2, d7, t62, zil131, and zsu23-4 from top to bottom.**

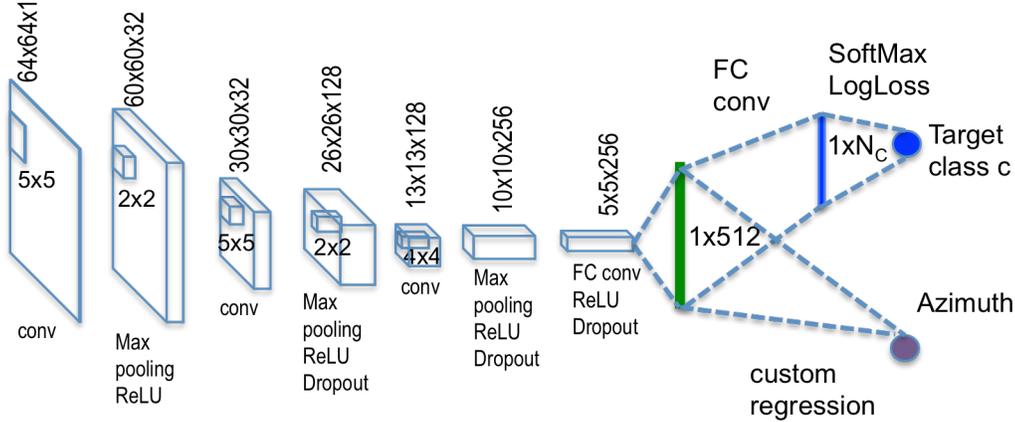

**Figure 3. Architecture of the pose-aware DCNN used for SAR ATR experiments on the public MSTAR dataset.**

## 6. Experiments

We have applied our pose-aware DCNN architecture to classify SAR chips on the publically available MSTAR SAR ATR dataset [36]. This dataset contains SAR chips for 10 target classes, collected at two or more depression angles, using an X-band SAR sensor at a 1-foot resolution spotlight mode. Each SAR image in this dataset is centered using its truth target position. Both training and test images cover the full azimuth range of $[0^o, 360^o]$. Figure 2 shows the targets and their sample SAR chips. Note that the azimuth angle of the target drastically affects the appearance, as different radar scattering mechanisms are illuminated as the target rotates relative to the sensor.

We have followed the convention to use all chips taken at $(17° \pm 1°)$ depression angle as the training set and those at $(15° \pm 1°)$ depression angle as the testing set [36]. This partition produces a total of 6,073 training images and 5,378 test images.

The public MSTAR dataset is rather small compared to training datasets typically used to train DCNNs. Out of concern for over-fitting, we designed a model of moderate size. We have used three convolutional layers followed by one fully connected layer, as shown in Figure 3.

For this SAR ATR problem all targets are on a ground plane. The 3D pose of a target degenerates to the azimuth angle of the target w.r.t the sensor. For this special case, Eq **2** of the distance between two pose becomes

**Eq 5:** $dist(q_1, q_2) = \arccos(|\cos((\theta_1^{azimuth} - \theta_2^{azimuth})/2)|)$

which has a range of $[0, \frac{\pi}{2}]$.

Noticing the symmetry in most of the targets (Figure 2) we have applied flip augmentation to expand the training dataset. We flip each training image left-right (in cross-range) and add the flipped image in the training set to double the size of the training set from 6,073 to 12,146. To train the pose-aware DCNN, we need to associate each flipped image with a pose. Since the azimuth angle is 0 when a target is head on, we negate the azimuth angle of the original training image and assign it to its flipped image. This flip augmentation has shown to improve the classification performance in our evaluations.

We train the model using mini-batch Stochastic Gradient Descent with momentum and weight decay. The weights and parameters of the network are randomly initialized with a zero mean Gaussian distribution with a standard deviation of 0.01. The other parameter values used in training are shown in Table 1.

**Table 1. Parameter values used in DCNN training**

| Parameter name | Value |
| --- | --- |
| Mini-batch size | 100 |
| Momentum | 0.9 |
| Weight decay | 0.0005 |
| Learning rate | 0.001 |
| Weight $\lambda$ for $obj_{pose}$ | 1.0 |

To assess the advantage of the pose-aware DCNN architecture, we also evaluate the performance of a baseline model, which is identical except that the pose constraint is removed during training, such that the performance difference between the two deep nets is solely due to the pose reasoning.

We show on the left and right side of Figure 4 the confusion matrix of the baseline network and the proposed pose-aware DCNN using the MSTAR dataset. Even though our baseline architecture has performed very well with an accuracy of 99.03% over all test images, the auxiliary pose information in training has sculptured the pose-aware DCNN to achieve an overall accuracy of 99.50%, which is an almost 50% relative error reduction.

Table 2 compares the performances of our proposed algorithm and existing approaches on the MSTAR dataset. Not surprisingly, the top three performances are achieved using DCNNs: our baseline model, A-ConvNets, and the proposed pose-aware model. Note that our baseline model performs quite on par with A-ConvNets [48], 99.03% versus 99.1% in classification accuracy, considering that [48] augmented the training set to 27,000 images, while we only have 12,146 training images after the flip-augmentation. The proposed pose-aware DCNN is able to improve upon the conventional architecture and achieve the best overall classification accuracy of 99.50% among currently published algorithms for the MSTAR dataset.

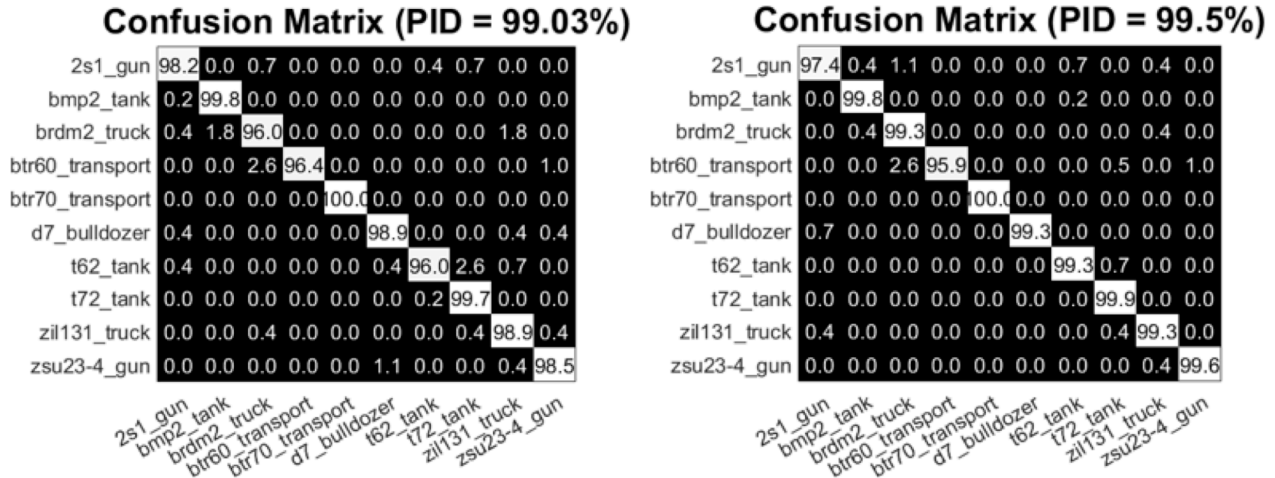

**Figure 4.** Confusion matrix (in percent) for SAR target classification using pose-aware DCNN (right) and baseline DCNN without pose information (left) on the public MSTAR dataset. Rows: truth; columns: result.

**Table 2. SAR ATR performance comparison with published algorithms.**

| Algorithm | Accuracy (%) |
| --- | --- |
| Support Vector Machines [55] | 90.1 |
| Adaboost [42] | 92.7 |
| Bayesian Compressive Sensing [53] | 92.6 |
| SRMS [8] | 93.6 |
| Conditionally Gaussian Model [32] | 96.9 |
| Modified Polar Mapping [33] | 98.8 |
| Basic DCNN [30] | 92.3 |
| A-ConvNets [48] | 99.1 |
| Proposed baseline DCNN | 99.0 |
| Proposed Pose-aware DCNN | 99.5 |

## 7. Conclusions

To take full advantage of deep neural networks we need to better understand how they work to solve highly complex real world challenges. Taking clues from observations that deep networks capture attributes or functionalities that do not directly associate with the tasks they are trained on, we perceive that deep networks build holistic and general representations in order to optimize an objective on a dataset full of variations from many different sources. Recognizing that deep nets perform unsupervised learning of impacting latent factors during supervised learning of a specific objective, we propose to boost training with available information on the auxiliary explanatory factors to obtain networks with better comprehension of the data population. We describe a general framework to incorporate knowledge of explanatory factors into the deep model for improved performance. We also apply this principle to train pose-aware DCNNs for classification and demonstrate that the auxiliary pose information helps improve the classification performance on a SAR ATR task.

As the world is full of variations and confounding factors are omnipresent for practical problems, our findings open up new possibilities to improve the performances of DCNNs. For example, it is possible to improve face identification by augmenting the training with age and gender information. We will explore applying this principle to additional deep learning applications with different explanatory factors.

## Acknowledgement

This research was developed with funding from the Defense Advanced Research Projects Agency (DARPA). The views, opinions and/or findings expressed are those of the authors and should not be interpreted as representing the official views or policies of the Department of Defense or the U.S. Government. Approved for Public Release, Distribution Unlimited.